\documentclass[letterpaper, 10 pt, conference]{ieeeconf}

\IEEEoverridecommandlockouts                              
\usepackage{graphics} 
\usepackage{epsfig} 
\usepackage{mathptmx} 
\usepackage{times} 
\usepackage{amsmath} 
\usepackage{amssymb}  
\usepackage{amssymb}  
\usepackage{multirow} 
\usepackage{booktabs}
\usepackage{adjustbox}  
\usepackage{pifont} 
\usepackage{tikz}
\usetikzlibrary{arrows.meta,positioning,calc,fit,decorations.pathreplacing}
\usepackage{subcaption}
\usepackage{pgfplots}
\pgfplotsset{compat=1.18}
\usetikzlibrary{intersections}
\usetikzlibrary{fillbetween}
\usepackage{booktabs,multirow}
\usepackage{array}
\usepackage[skip=4pt]{caption}

\title{\LARGE \bf
Adaptive Capacity Allocation for Vision Language Action Fine-tuning
}

\author{
Donghoon Kim$^{1}$, Minji Bae$^{1\dagger}$, Unghui Nam$^{1\dagger}$, Gyeonghun Kim$^{1\dagger}$, Suyun Lee$^{1\dagger}$,\\
Kyuhong Shim$^{2\ddagger}$, Byonghyo Shim$^{1\ddagger}$
\thanks{This work was supported by the National Research Foundation of Korea(NRF) grant funded by the Korea government(MSIT) (2022M3C1A3099336) and in part by Samsung Electronics Co., Ltd (IO251211-14335-01).}
\thanks{$^{1}$D.Kim, M.Bae, U.Nam, G.Kim, S.Lee, B.Shim are with Department of Electrical and Computer Engineering, Seoul National University, Seoul 08826, Republic of Korea (email: $\{$dhkim, mjbae, uhnam, ghkim, sylee, bshim$\}$@islab.snu.ac.kr}
\thanks{$^{2}$K.Shim is with Department of Computer Science and Engineering, Sungkyunkwan University, Suwon 16419, Republic of Korea (email: khshim@skku.edu)}
\thanks{$^{\dagger}$2nd co-authors, listing order is random}
\thanks{$^{\ddagger}$Corresponding authors}
}

\begin{document}

\maketitle
\thispagestyle{empty}
\pagestyle{empty}

\begin{abstract}
Vision language action models (VLAs) are increasingly used for Physical AI, but deploying a pre-trained VLA model to unseen environments, embodiments, or tasks still requires adaptation.
Parameter-efficient fine-tuning (PEFT), especially LoRA, is common for VLA policies, yet the exposed capacity knob, the rank, does not transfer uniformly: robotics transfer exhibits a higher and task-varying intrinsic rank than language fine-tuning. 
Small ranks suffice for LLMs (e.g., $r\!\in\!\{4,8\}$), while spectral analyses indicate VLAs may require much larger ranks (e.g., $r\!\approx\!128$) or near–full rank, a mismatch that worsens in multi-task settings.
We present LoRA-SP (Select–Prune), a rank-adaptive fine-tuning method that replaces fixed-rank updates with input- and layer-wise capacity. LoRA-SP uses an SVD-style parameterization with a small router whose nonnegative scores act as singular values over a shared vector bank. The active set is chosen by an energy target on the cumulative squared scores \(E(k)\ge \eta\), providing a direct link to approximation error via our spectral analysis. During training, \(\eta\) concentrates energy on a few directions and teaches the router to rely on fewer vectors while preserving accuracy. This yields compact adapters that reduce cross-task interference and improve generalization.
On four real-robot manipulation tasks collected on an unseen AgileX PiPER arm, across two VLA backbones ($\pi_{0}$ and SmolVLA), LoRA-SP matches or exceeds full fine-tuning with far fewer trainable parameters, and improves multi-task success by up to 31.6\% over standard LoRA while remaining robust to rank choice.
\end{abstract}

\section{INTRODUCTION}
In recent years, large multimodal models (LMMs) have been applied in a wide range of domains long considered exclusive to human intelligence, such as complex reasoning, tool use, and code generation~\cite{openai2023gpt4,gemini2023,llava2023}.
Building on this progress, LMMs are moving beyond image–text perception to physical interaction with the world~\cite{brohan2022rt1,brohan2023rt2,kim2024openvla,deepmind2025geminirobotics}. 
This shift has given rise to physical AI: building agents that act and learn through embodied interaction in real-world settings.
 
Conventionally, approaches based on reinforcement learning (RL) and imitation learning (IL) have been widely used to build embodied intelligence, but these approaches are often limited to specific tasks or environments~\cite{levine2016e2e,rajeswaran2018dexterous,kalashnikov2018qtopt}.
Recent progress has been driven by the training of LMMs with vision-language-action-paired data, enabling agents to learn generalizable mappings from visual perception and language instruction to action (a.k.a., vision-language-action (VLA) model)~\cite{oxe2023rtx,black2024pi0}.
This paradigm allows a single model to operate in diverse environments, embodiments, and task distributions, marking a significant step toward versatile embodied intelligence.

\begin{figure*}[t!]
\centering
\includegraphics[width=\textwidth]{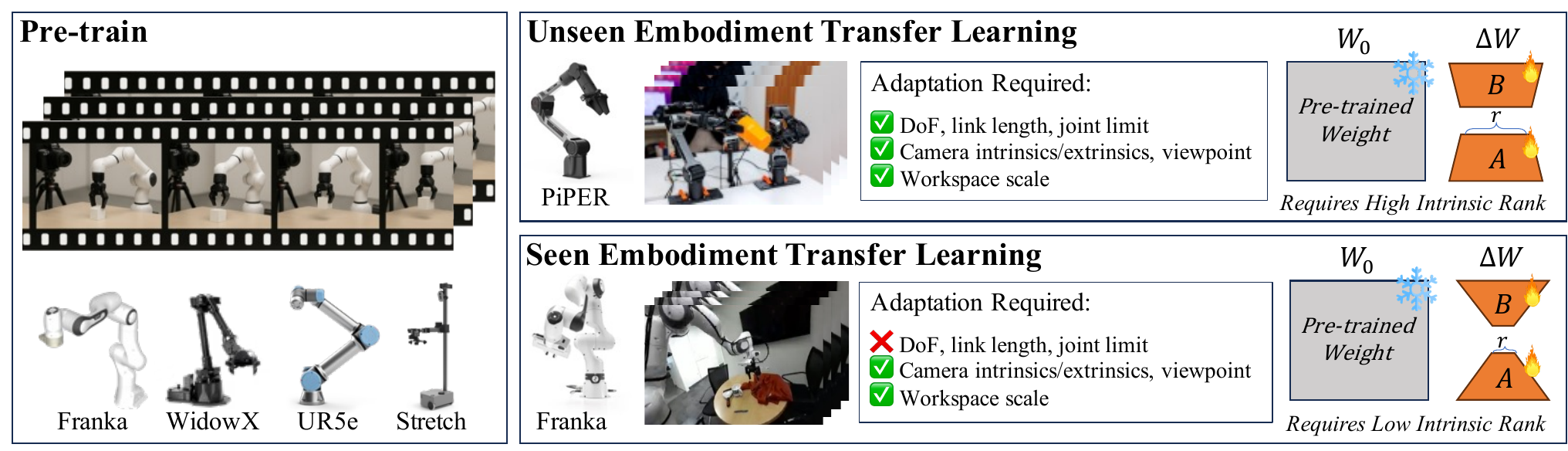}
\caption{
VLA models are pre-trained on diverse manipulation tasks and robot embodiment (e.g., Franka Emika, WidowX, UR5e, Stretch) data. 
We compare transfers to a \emph{seen} embodiment (Franka) versus an \emph{unseen} embodiment (PiPER). 
Unseen-embodiment transfer changes both the robot’s kinematic specification (DoF, link lengths, joint limits) and the perception geometry (camera intrinsics/extrinsics, viewpoint) and workspace scale, which pushes the update to require a \emph{higher intrinsic rank}; by contrast, when the embodiment is seen, adaptation primarily compensates for perception/scale shifts and works with a \emph{lower rank}.}
\label{fig:overview}
\vspace{-0.5cm}
\end{figure*}

Although Vision-Language-Action (VLA) models have made significant strides, the idea of a single universal model remains challenging. 
This is especially true when the robot's embodiment, task, or environment differs from those in the training data. 
As shown in Fig.~\ref{fig:overview}, differences in hardware specifications (e.g., DoF, link lengths, and joint limits) change the feasible inverse kinematics set, so identical goals map to different joint-space solutions and thus different trajectories.
Shifts in perception–action alignment, driven by camera intrinsics/extrinsics, viewpoint, and workspace scale, also change how pixel displacements translate into physical motion in the robot frame.
In summary, these real-world factors modify the distribution that the policy needs to cover, thereby increasing the adaptation capacity required.  

\begin{figure}[t!]
\centering
\includegraphics[width=0.8\linewidth]{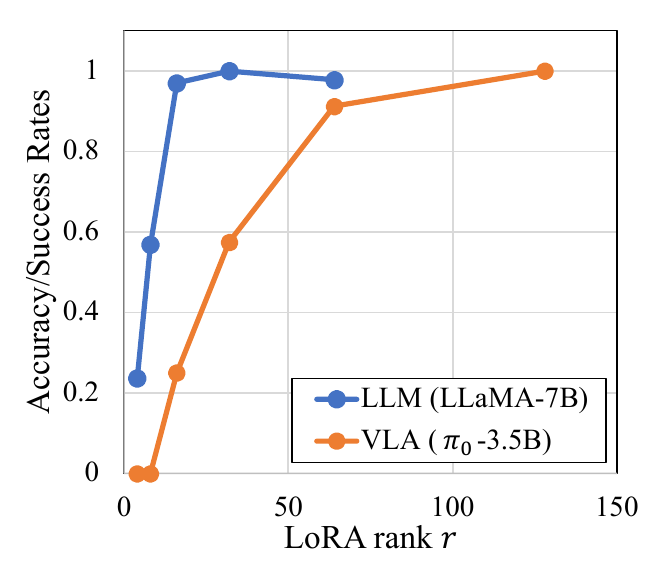}
\caption{Rank--performance curves (accuracy/success relative to full fine-tuning; 1.0 = full FT). 
LLM (LLaMA-7B) reaches near-full-FT performance with very small ranks ($r\in\{4,8\}$), 
whereas VLA ($\pi_0$-3.5B) improves steadily and only approaches parity around $r\approx128$, 
consistent with a higher intrinsic dimension in the VLA transfer setting.}
\label{fig:rank_performance}
\vspace{-0.5cm}
\end{figure}
In this adaptation task, LoRA and its variants have been widely used due to their competitive performance and high parameter efficiency~\cite{Hu2022LoRA,dettmers2023qlora,liu2024dora}.
However, the rank hyperparameter, which is key to their capacity, does not generalize uniformly across different domains, meaning that a fixed rank might not be uniformly optimal for all cases.
We observe that, for adaptation in robotics learning, the intrinsic rank (i.e., the minimal dimensionality required to capture task-relevant features) is typically higher and more variable compared to language models~\cite{Li2018IntrinsicDimension,Aghajanyan2021IntrinsicDim}.
As shown in Fig~\ref{fig:rank_performance}, while LLaMA-7B achieves near full fine-tuning performance with rank $r\in\{4,8\}$, ${\pi_0}$-3.5B demands ranks up to $r=128$ to achieve the same performance~\cite{touvron2023llama, black2024pi0}.
Even in single-task training, the optimal rank varies with task difficulty (see Fig.~\ref{fig:rank_performance_single_multi} (a)).
This uncertainty about the rank makes it difficult to select a single global rank in multi-task settings (see Fig.~\ref{fig:rank_performance_single_multi} (b)).
Such a choice may force heterogeneous tasks to share a fixed subspace, which increases cross-task interference (competition for the shared adapter subspace) and reduces positive transfer.
In practice, we perform the brute sweeping (e.g., grid search) of the rank parameter to find a near-optimal rank for each setting.
The cost of brute-force sweeps highlights the need for rank-adaptive frameworks that dynamically allocate capacity to task- and embodiment-specific demands.

To this end, we introduce LoRA-SP (Select-Prune), a rank-adaptive fine-tuning technique designed to resolve the limitation of fixed-rank LoRA approaches.
Classic LoRA updates the weight $W$ with a fixed-rank factorization $\Delta W = BA$ so the module computes $(W+\Delta W)x = Wx + B(Ax)$.
We generalize this low-rank form by replacing $BA$ with SVD-style parameterization $U\,\mathrm{diag}(s(x))\,V$ where $U$ and $V$ define a vector bank (basis) and the router outputs nonnegative \textit{singular-value-like} scores $s(x)\in\mathbb{R}^r$.
We start from a sufficiently large rank $r$ (smaller than dense layers but large enough to cover necessary directions) and let the router learn, per input and per layer, which basis vectors are active and what magnitudes (singular values) they should take.
We then choose the smallest active rank $k$ such that the cumulative singular-value energy satisfies $E(k)\ge\eta$, zeroing the remaining vectors for that input.
We also add a spectral loss $\mathcal{L}_{\text{spec}}=1-E_k(x)$ that concentrates energy onto the selected vectors, creating a progressive concentration where useful directions are amplified across iterations.
As a result, LoRA-SP learns to realize tasks with only the minimal active set needed, yielding compact adapters that preserve accuracy while lowering inference cost.

We evaluate our method on four real-world manipulation tasks collected with an unseen 7-DoF AgileX PiPER arm, totaling 480 demonstrations with dual RGB views. Experiments are conducted on two pretrained VLA backbones: $\pi_0$\cite{black2024pi0} and SmolVLA\cite{shukor2025smolvlavisionlanguageactionmodelaffordable}, which represent a high-capacity and a lightweight model, respectively. These are compared against several baselines including full fine-tuning, standard LoRA with ranks $r \in \{16, 32, 64, 128\}$, AdaLoRA~\cite{Zhang2023AdaLoRA}, and LoRA-MoE~\cite{dou2024loramoe} using top-1 and weighted-sum routing.
Standard LoRA shows clear rank sensitivity, performing well in single-task settings at high ranks but collapsing under multi-task training due to mismatched task capacities and subspace interference. Neither AdaLoRA nor LoRA-MoE consistently outperform standard LoRA in the multi-task regime. In contrast, LoRA-SP achieves consistently strong multi-task performance across all tasks and backbones, while updating significantly fewer parameters than full fine-tuning (Table~\ref{tab:main}, \ref{tab:main_lora}).
An ablation study shows that the spectral loss enables effective rank pruning without accuracy loss, while threshold ablation confirms LoRA-SP’s robustness under reduced active ranks (Table~\ref{tab:ablation_loss}, \ref{tab:ablation_threshold}).

Our contributions are summarized as follows:

\begin{itemize}
    \item 
    We quantify rank needs via cumulative energy $E(k)$ and rank–performance curves, showing that OOD embodiment transfer (e.g., AgileX PiPER) requires substantially larger ranks than language fine-tuning and exhibits strong rank sensitivity (Figs.~\ref{fig:rank_performance}, \ref{fig:overview}). 
    This motivates rank-adaptive capacity instead of a fixed global rank.

    \item 
    We introduce a fine-tuning method that adaptively adjusts trainable capacity per input and layer.
    Router produces singular-value–like scores $s(x)$ over a shared vector bank, and the effective rank is set by an energy target on the cumulative squared scores. 
    Spectral concentration loss amplifies the surviving vectors, creating a positive feedback that progressively reduces the active set, while the task loss preserves accuracy.

    \item 
    We validate LoRA-SP on four real-world manipulation tasks with 7-DoF AgileX PiPER arm, using $\pi_0$ and SmolVLA backbones. Compared to the baselines LoRA-SP achieves comparable or better performance with significantly fewer trainable parameters and activated ranks. It improves multi-task success by up to 31.6\% over  standard LoRA while remaining robust to rank choice (Sec.~\ref{sec:experiments}, Tabs.~\ref{tab:main}, Tabs.~\ref{tab:main_lora},\ref{tab:ablation_loss}).
\end{itemize}

\section{Theoretical Background and Rationale}
\subsection{Intrinsic Dimension of Fine-tuning}
We formalize the \emph{intrinsic dimension} (ID) of a task as the smallest update capacity needed to recover a target performance (e.g., the performance achieved by full fine-tuning). Let a layer have parameters \(W\in\mathbb{R}^{d_{\mathrm{out}}\times d_{\mathrm{in}}}\), training objective \(L(\cdot)\), initialization \(W_0\), and a target value \(L_\star\) (e.g., the loss achieved by full fine-tuning). Under LoRA, we constrain the update to be low rank, \(\Delta W = BA\), with \(B\in\mathbb{R}^{d_{\mathrm{out}}\times r}\) and \(A\in\mathbb{R}^{r\times d_{\mathrm{in}}}\), so that \(\mathrm{rank}(\Delta W)\le r\) \cite{Li2018IntrinsicDimension,Hu2022LoRA}. For tolerance \(\varepsilon\ge 0\), we define the LoRA-based intrinsic dimension as
\begin{equation}
\label{eq:lora-id}
\mathrm{ID}^{\mathrm{LoRA}}_\varepsilon
=\min\left\{\, r:\;
\begin{array}{l}
\exists\,A\in\mathbb{R}^{r\times d_{\mathrm{in}}},\ 
B\in\mathbb{R}^{d_{\mathrm{out}}\times r}\\[2pt]
\text{s.t. } L\big(W_0 + BA\big) \le L_\star + \varepsilon
\end{array}
\right\}.
\end{equation}
In words, instead of optimizing all entries of \(W\), we restrict the update to a rank-\(r\) factorization \(BA\). If such \((A,B)\) achieve the target loss within \(\varepsilon\), then \(r\) is feasible; \(\mathrm{ID}^{\mathrm{LoRA}}_\varepsilon\) is the minimum feasible rank. This yields an operational measure of the task’s effective degrees of freedom that is largely decoupled from the ambient parameter count.

Building on this structure, subsequent methods tie rank selection to data- and layer-specific sensitivity using spectral diagnostics. 
For example, AdaLoRA allocates a global rank budget across layers using SVD-based importance scores, placing the capacity where it is needed most \cite{Zhang2023AdaLoRA}.

\begin{figure}[t!]
\centering
\includegraphics[width=\linewidth]{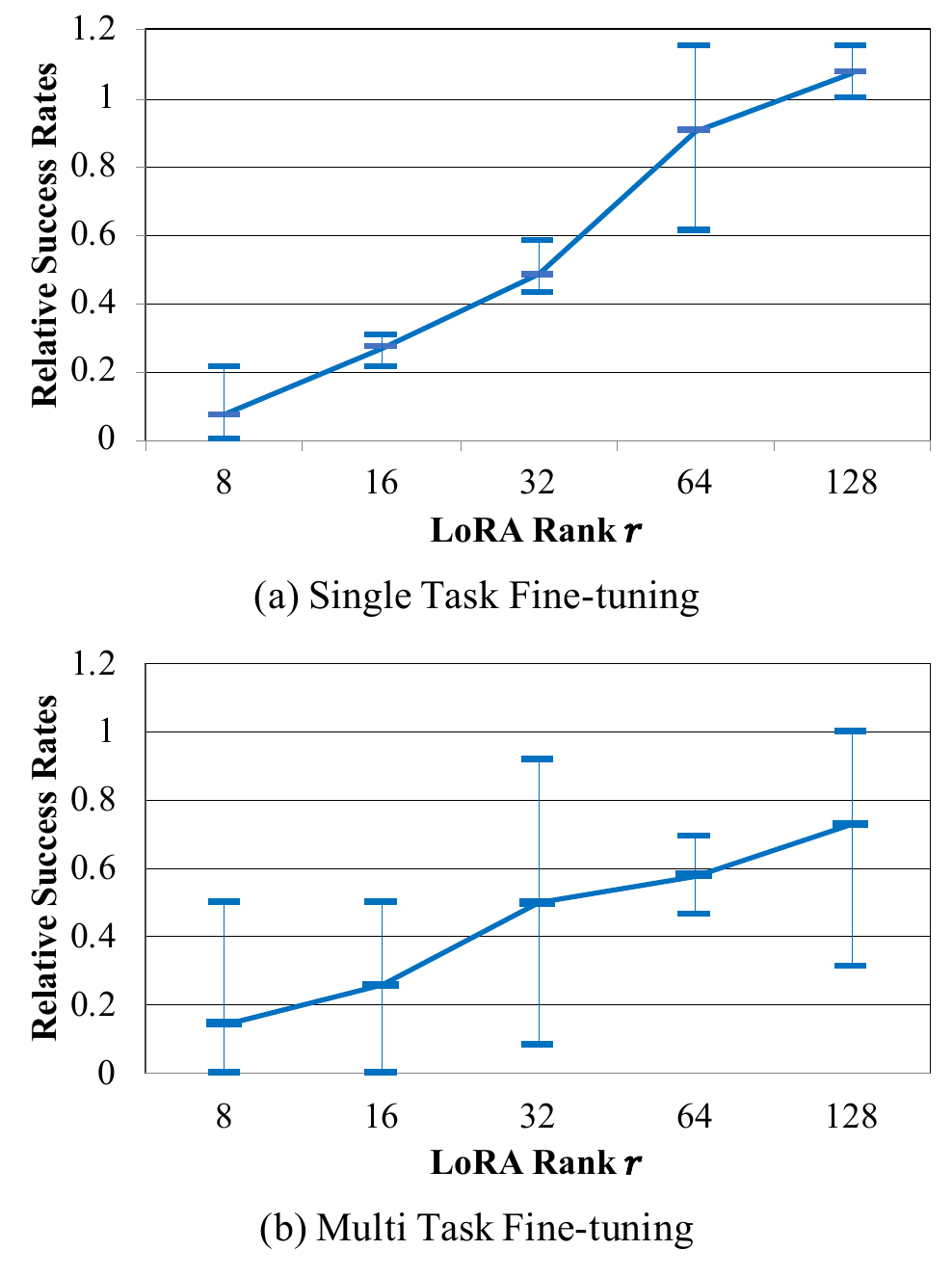}
\caption{{Rank sensitivity in single- and multi-task LoRA fine-tuning on $\pi_0$ model.}
(a) LoRA modules trained independently on each single task. While single-task modules also require higher ranks to reach full performance, their variance across tasks is lower than in the multi-task setting. Together, the results highlight the difficulty of choosing a single global rank that balances efficiency and accuracy across tasks, motivating rank-adaptive allocation.
(b) Multi-task LoRA fine-tuning across four manipulation tasks. Success rate increases with rank but exhibits substantial variance across tasks, reflecting interference and heterogeneous capacity needs. }

\label{fig:rank_performance_single_multi}
\vspace{-0.5cm}
\end{figure}
\subsection{LoRA-MoE and Multi-task Performance}
A common extension of LoRA to multi-task settings is to view each task as a small low-rank perturbation applied to a shared base policy. 
Let $f(x;\theta)$ be the frozen base model and, for each task $i$ (e.g., pick-and-place, tool use), let $\Delta\theta_i$ be a LoRA update such that the single-task policy $f_i(x)=f(x;\theta+\Delta\theta_i)$ performs well on task $i$. 
In a pure single-task fine-tune we would deploy $\theta+\Delta\theta_i$ for that task; in the multi-task case we instead add a routing function $g(x)$ that selects or mixes these adapters at inference, yielding
\begin{equation}
\tilde{f}(x)=f\!\left(x;\,\theta+\sum_i g_i(x)\,\Delta\theta_i\right),
\label{eq:LoRA-MoE}
\end{equation}
where $g(x)$ can be a hard selector ($g_i\in\{0,1\}$, one expert per input) or a soft mixture ($\sum_i g_i=1$)~\cite{dou2024loramoe}. 
Under ideal routing and disjoint task distributions, $\tilde{f}$ matches each single-task expert on its own data. 
In practice, however, capacity scales with the number of experts, increasing memory and latency; performance is also sensitive to routing errors; and useful update directions remain siloed within task-specific adapters, limiting sharing.
Moreover, both the LoRA rank per expert and the number of experts are sensitive hyperparameters that typically require dataset-specific sweeps. 
These factors make LoRA-MoE hard to tune and deploy at scale; by contrast, LoRA-SP uses a single adapter with input- and layer-conditioned capacity, avoiding expert proliferation while enabling parameter sharing (see Sec.~\ref{sec:methods}).

\begin{figure*}[t!]
\centering
\includegraphics[width=1.01\textwidth]{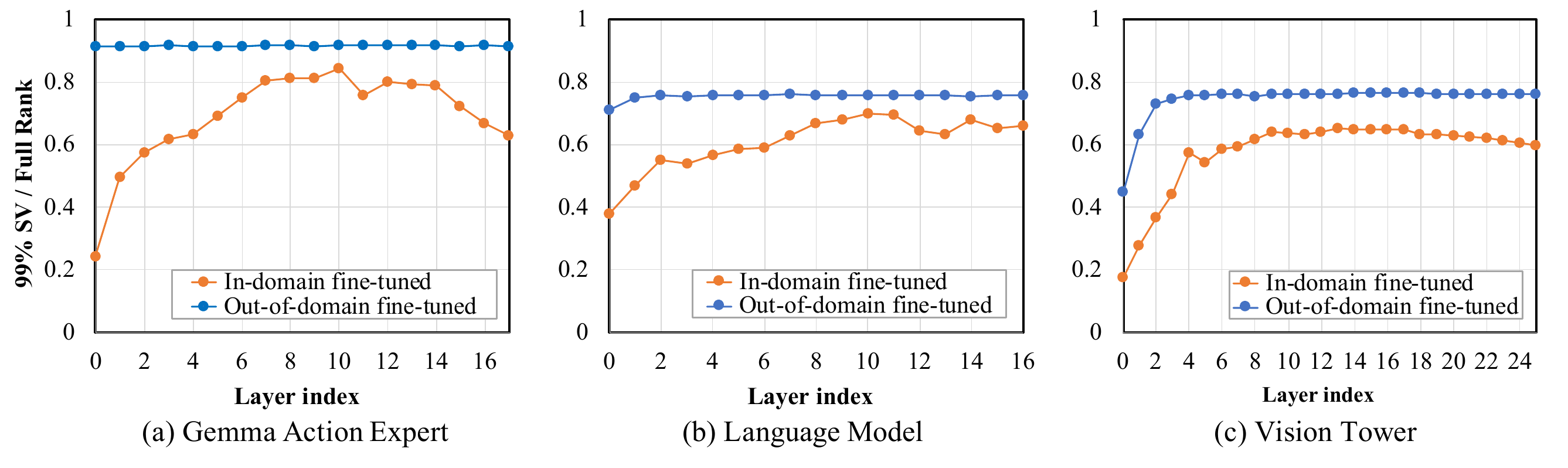}
\caption{\textbf{Spectral Rank Variation by Embodiment During $\pi_0$ Fine-tuning.}
Number of singular values required to capture 99\% of the total energy (normalized by the full rank) across different layers and modules. We compare $\pi_0$ models fine-tuned on in-domain and out-of-domain data. The in-domain model is fine-tuned on the DROID dataset, which uses the robotic arm (Franka Panda) included in $\pi_0$'s pretraining data. The out-of-domain model is fine-tuned on a dataset collected with the AgileX PiPER robotic arm, an embodiment absent from the pretraining data. The results show that the required rank varies by embodiment, and generalizing to a novel embodiment demands higher-rank to achieve comparable performance.}

\label{fig:energy-curves}
\vspace{-0.5cm}
\end{figure*}
\section{Analysis: Intrinsic Dimension and Spectral Error}
\label{sec:analysis}

\subsection{Spectral Error in Low Rank Approximation of Gradients}
Given a rank $r$ on a weight update, \emph{the smallest attainable relative Frobenius error is determined by the spectrum}:
if $E(k)$ denotes the cumulative energy of the top-$k$ singular values (Eq.~\ref{eq:energy}), then the \emph{best} rank-$k$ approximation
achieves error $\sqrt{1-E(k)}$ (Eq.~\ref{eq:relerr}). Thus, \emph{choosing ranks} in LoRA is equivalent to \emph{controlling}
these spectral error.

\vspace{3pt}
\noindent
\textbf{Definition}: Let $A$ denote a weight or update matrix. Write its singular values as
$\sigma_1\!\ge\!\cdots\!\ge\!\sigma_r\!>\!0$ with $r=\mathrm{rank}(A)$, and define the cumulative energy
\begin{equation}
\label{eq:energy}
E(k)\;=\;\frac{\sum_{i=1}^{k}\sigma_i^2}{\sum_{i=1}^{r}\sigma_i^2}\,.
\end{equation}
If $A_k$ is the rank-$k$ truncated SVD of $A$, then we have the following identity and optimality statement.

\vspace{3pt}
\noindent
\textbf{Proposition}:
Let $A=U\Sigma V^\top$ be an SVD parametrization with $\Sigma=\mathrm{diag}(\sigma_1,\ldots,\sigma_r,0,\ldots)$
and let $\Sigma_k=\mathrm{diag}(\sigma_1,\ldots,\sigma_k,0,\ldots)$ so that $A_k=U\Sigma_k V^\top$.
Then
\begin{equation}
\label{eq:relerr}
\frac{\|A-A_k\|_F}{\|A\|_F}\;=\;\sqrt{\,1-E(k)\,}\,,
\end{equation}
equivalently,
\begin{equation}
\label{eq:relerr2}
\|A-A_k\|_F^2=\sum_{i=k+1}^r \sigma_i^2.
\end{equation}
Moreover, $A_k$ is a best rank-$k$ approximation to $A$ in Frobenius norm, i.e.,
$\|A-B\|_F\ge \|A-A_k\|_F$ for every matrix $B$ with $\mathrm{rank}(B)\le k$.

\vspace{3pt}
\noindent
\textbf{Proof}:
By orthogonal invariance of the Frobenius norm,
\begin{equation}
\|A-A_k\|_F
=\|U(\Sigma-\Sigma_k)V^\top\|_F
=\|\Sigma-\Sigma_k\|_F.
\end{equation}
Since $\Sigma-\Sigma_k=\mathrm{diag}(0,\ldots,0,\sigma_{k+1},\ldots,\sigma_r)$, we obtain
\begin{equation}
\|A-A_k\|_F^2
=\|\Sigma-\Sigma_k\|_F^2
=\sum_{i=k+1}^r \sigma_i^2.
\end{equation}
Also,
\begin{equation}
\|A\|_F^2
=\|\Sigma\|_F^2
=\sum_{i=1}^r \sigma_i^2.
\end{equation}
Dividing the two identities yields Eq.~\eqref{eq:relerr} with $E(k)$ as in Eq.~\eqref{eq:energy}.
The optimality of $A_k$ among all rank-$k$ matrices follows from the Eckart–Young–Mirsky theorem.  
\hfill$\square$

\subsection{Characteristics of Gradients in VLA Models}

\begin{figure*}[t!]
\centering
\includegraphics[width=\textwidth]{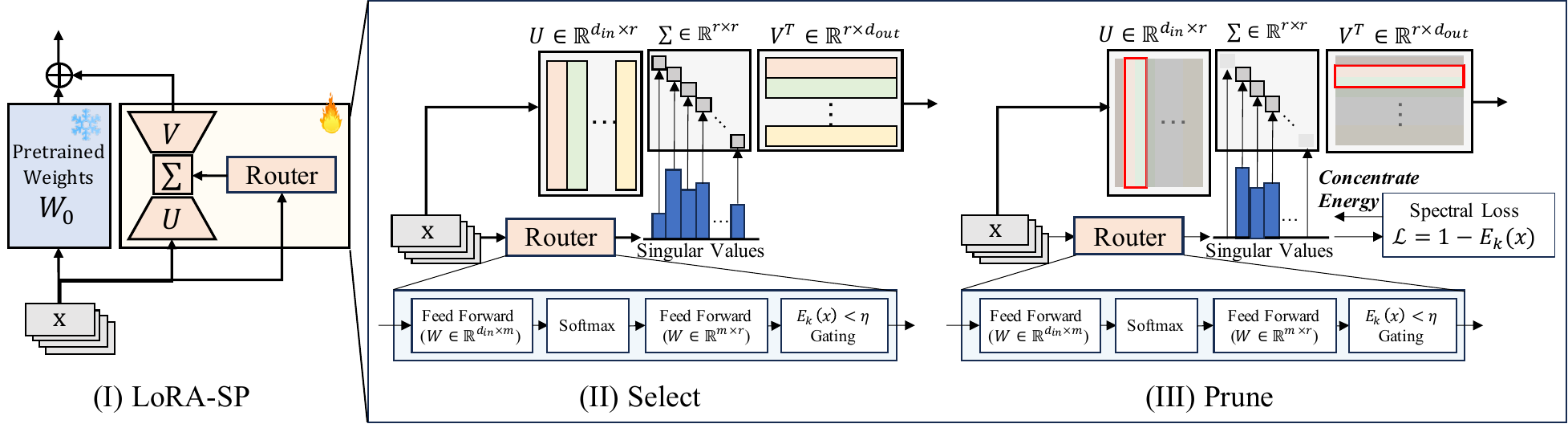}
\caption{\textbf{LoRA-SP (Select--Prune).}
(I) Overview: a wide vector bank $(U,V)$ is trained together with a router on the backbone $W_0$.
(II) Select: the router produces vector-level scores that act as singular values, forming an input- and layer-conditioned update $\Delta W = U\,\Sigma(x)\,V$; the histogram illustrates the spectral energy distribution across vectors.
(III) Prune: only the smallest set of basis vectors whose cumulative energy exceeds the target $\eta$ are kept, progressively reducing the active rank while maintaining accuracy.
}
\label{fig:method}
\vspace{-0.5cm}
\end{figure*}



\noindent
\textbf{Intrinsic Dimension}
As shown in the rank–performance curves of Fig.~\ref{fig:rank_performance}, while the language model (LLaMA-7B) approaches full fine-tuning (FT) performance even at low ranks ($r\in\{4,8\}$), the vision–language–action (VLA) model $\pi_0$-3.5B requires ranks up to $r\approx128$ to achieve full FT performance. 
This suggests that the intrinsic dimension of the gradient in VLA tasks is significantly higher, necessitating more expressive update directions for effective adaptation. 
Supporting evidence from the spectral energy curves in Fig.~\ref{fig:energy-curves} reveals that the minimum rank $k$ satisfying $E(k)\ge0.99$ (based on the relative error bound in Eq.~\eqref{eq:relerr}) varies widely between modules and downstream dataset, ranging from $0.2$ to $0.9$ of the full rank in the Gemma expert, language model, and vision tower. 
This wide distribution of required ranks across layers and domains clearly exposes the limitations of uniform rank allocation.

\vspace{3pt}
\noindent
\textbf{Rank Sensitivity}  
Fine-tuning $\pi_0$ in out-of-domain settings (e.g., robotic arms not seen during pre-training) further increases the effective dimension of gradients. 
For instance, Fig.~\ref{fig:energy-curves} shows that datasets collected with the AgileX PiPER robotic arm (out-of-domain) consistently require higher spectral ranks across all modules compared to in-domain data from the Franka Panda. 
Similar patterns emerge in multi-task learning (Fig.~\ref{fig:rank_performance_single_multi}): while single-task modules also demand high ranks, their performance variance across tasks is lower than in multi-task settings. 
When fine-tuning across four manipulation tasks, substantial performance variance arises due to task interference and heterogeneous capacity demands. 
This highlights the difficulty of balancing efficiency and accuracy with a single global rank, reinforcing the need for adaptive rank allocation.

\section{LoRA Select--Prune (LoRA-SP)}\label{sec:methods}

\subsection{Overview}\label{sec:overview}
LoRA-SP replaces a single fixed rank with data-conditioned capacity that varies by input and layer. 
As shown in Fig.~\ref{fig:method}, we train a shared vector bank $(U,V)$ and a small router. 
For each input $x$, the router scores basis vectors and we keep only the smallest set whose cumulative score energy meets a target $\eta$. 
This realizes the update with a few active vectors at test time and reduces interference across tasks by reusing only the vectors that matter for the current input.
\subsection{Problem Setup and Connection to LoRA}\label{sec:setup}
We adapt a pretrained backbone $f(W;\,x)$ with layer weights $\{W_\ell\}$ for multi-task data $(x,t)$.
Classic LoRA applies a fixed-rank update $\Delta W = BA$ so the forward map is $(W+\Delta W)x = Wx + B(Ax)$.
LoRA-SP generalizes this by replacing $BA$ with an input-conditioned SVD-style form
\begin{equation}
\Delta W_\ell(x) \;=\; U_\ell\,\mathrm{diag}\!\big(s_\ell(x)\big)\,V_\ell,
\end{equation}
where $U_\ell\in\mathbb{R}^{d_{\text{out}}\times r}$ and $V_\ell\in\mathbb{R}^{r\times d_{\text{in}}}$ define a per-layer vector bank, and $s_\ell(x)\in\mathbb{R}^r_{\ge 0}$ indicates router scores. 
Intuitively, $s_\ell(x)$ plays the role of data-conditioned singular values: large scores mark directions that should be active for the current input.

\subsection{Select: Vector-level Gating}\label{sec:select}
\vspace{3pt}
\noindent\textbf{Wide Initialization}
We initialize a wide vector bank $U$ and $V$ per module ($r{=}128$ in all experiments) and train it jointly with a lightweight two-layer router (Fig.~\ref{fig:method} (II)). 
Although full rank would be ideal in principle, before adaptation we do not know which directions in the representation space will be useful for a given input or layer. 
A wide initialization therefore provides sufficient coverage of candidate directions without committing to a specific subspace; the router and the spectral loss then discover a compact, task-relevant subspace and deactivate the rest.

\vspace{3pt}
\noindent\textbf{Singular Value Generation} 
Unlike module-level MoE (routes to one expert per call), LoRA-SP gates at the vector level, so the number of active vectors (i.e., LoRA rank) is adjusted per input and per layer.
Given input $x\in\mathbb{R}^{d_{\text{in}}}$, the router produces singular value-like scores $s(x)$ through a two-layer MLP with activation $\phi$:
\begin{equation}
h_1(x)=\phi(W_1x+b_1), \qquad s(x)=W_2h_1(x)+b_2\in\mathbb{R}^r_{\ge 0}.
\end{equation}
With $U\in\mathbb{R}^{d_{\text{out}}\times r}$ and $V\in\mathbb{R}^{r\times d_{\text{in}}}$, the update becomes
\begin{equation}
\Sigma(x)=\mathrm{diag}\big(s(x)\big), \qquad \Delta W(x)=U\,\Sigma(x)\,V.
\end{equation}
This construction lets LoRA-SP flexibly adjust both the rank and the direction of updates.

\begin{table*}[t]
\caption{Trainable parameters, active ranks, and multi-task success rates across various fine-tuning strategies. Active rank denotes effective rank $k$ per token in average across all layers.}
\centering
\small                              
\setlength{\tabcolsep}{5pt}      
\def\arraystretch{1.05}             
\setlength\heavyrulewidth{1.2pt}     
\begin{adjustbox}{width=0.85\textwidth}
\begin{tabular}{c l >{\centering\arraybackslash}m{1.5cm} >{\centering\arraybackslash}m{1.5cm} cccc}
\toprule
\multirow{2}{*}{Model} & \multirow{2}{*}{Strategy} 
& \multirow{2}{*}{\shortstack{\rule{0pt}{2.5ex}Trainable \\ / Total (\%)}} 
& \multirow{2}{*}{\shortstack{\rule{0pt}{2.5ex}Active \\ Rank}} 
& \multicolumn{4}{c}{Multi-task Success Rate (\%)} \\
\cmidrule(l){5-8}
                       &                           &                           &                         & Open & Pour & Press & Pick-Place \\
\midrule
\midrule

\multirow{6}{*}{$\pi_{0}$ ~\cite{black2024pi0}}     
                       & LoRA ($r=128$) ~\cite{Hu2022LoRA}            & 9.1 & 128 & 73.3 & 26.7 & 80.0 & 60.0   \\
                       & LoRA-MoE (top-1) ~\cite{dou2024loramoe}         & 9.2 & 32  & 13.3 & 13.3 & 53.3 & 13.3  \\
                       & LoRA-MoE (weighted sum) ~\cite{dou2024loramoe}   & 9.2 & 128 & 46.7 & 60.0 & 93.3 & 80.0   \\  
                       & AdaLoRA  ~\cite{Zhang2023AdaLoRA}                 & 9.1 & 76 & 20.0 & 6.7 & 40.0 & 60.0   \\ 
                       & Full FT                   & 100.0 & Full & 80.0 & 86.7 & 80.0 & 86.7   \\
                       & \textbf{LoRA-SP}         & 9.2 & 76 & 80.0 & 80.0 & 93.3 & 80.0  \\
\midrule

\multirow{6}{*}{SmolVLA ~\cite{shukor2025smolvlavisionlanguageactionmodelaffordable}} 
                       & LoRA ($r=128$) ~\cite{Hu2022LoRA}           & 17.0  & 128 & 40.0 & 20.0 & 93.3 & 86.7   \\
                       & LoRA-MoE (top-1) ~\cite{dou2024loramoe}         & 17.2 & 32 & 33.3 & 46.7 & 86.7  & 73.3  \\
                       & LoRA-MoE (weighted sum) ~\cite{dou2024loramoe}  & 17.2 & 128 & 60.0 & 80.0 & 100.0  & 66.7 \\     & AdaLoRA  ~\cite{Zhang2023AdaLoRA}                 & 17.0 & 60 & 6.7 & 0.0 & 40.0 & 20.0   \\    
                       & Full FT                   & 100.0 & Full & 73.3 & 86.7 & 100.0 & 86.7   \\
                       & \textbf{LoRA-SP}         & 17.1 & 60 &  86.7 & 86.7  &  100.0 & 93.3   \\
\bottomrule
\end{tabular}
\end{adjustbox}
\label{tab:main}
\end{table*}

\begin{table*}[t]
\caption{Comparison of task success rates across varying LoRA ranks under single-task and multi-task training.}
\centering
\small                              
\setlength{\tabcolsep}{3pt}      
\def\arraystretch{1.05}             
\setlength\heavyrulewidth{1.2pt}     
\begin{adjustbox}{width=0.85\textwidth}
\begin{tabular}{c l >{\centering\arraybackslash}m{1.5cm} >{\centering\arraybackslash}m{1.5cm} cccccccc}
\toprule
\multirow{3}{*}{Model} & \multirow{3}{*}{Strategy} 
& \multirow{3}{*}{\shortstack{\rule{0pt}{2.5ex}Trainable \\ / Total (\%)}} 
& \multirow{3}{*}{\shortstack{\rule{0pt}{2.5ex}Active \\ Rank}} 
& \multicolumn{8}{c}{Success Rate (\%)} \\
\cmidrule(l){5-12}
                       &                           &                           &                         & \multicolumn{4}{c}{Single-task Training} & \multicolumn{4}{c}{Multi-task Training} \\
\cmidrule(l){5-8}
\cmidrule(l){9-12}
                       &                           &                           &                         & Open & Pour & Press & Pick-Place & Open & Pour & Press & Pick-Place \\
\midrule
\midrule

\multirow{7}{*}{$\pi_{0}$ ~\cite{black2024pi0} }  
                       & \multirow{5}{*}{LoRA  ~\cite{Hu2022LoRA}}   & 0.6 & 8    & 6.7  & 0.0  & 0.0   & 20.0   & 0.0   &  6.7  & 40.0  & 0.0        \\
                       &   & 1.2 & 16       & 26.7 & 20.0 & 26.7 & 20.0 & 0.0 & 20.0 & 40.0 & 26.7   \\
                       &   & 2.4 & 32       & 40.0 & 46.7 & 40.0 & 40.0 & 6.7 & 53.3 & 73.3 & 33.3   \\
                       &   & 4.8 & 64       & 53.3 & 73.3 & 100.0 & 86.7 & 46.7 & 40.0 & 46.7 & 60.0   \\
                       &   & 9.1 & 128      & 93.3 & 80.0 & 100.0 & 100.0 & 73.3 & 26.7 & 80.0 & 60.0   \\
                       & Full FT                   & 100.0  & Full & 86.7 & 80.0 & 86.7 & 93.3 & 80.0 & 86.7 & 80.0 & 86.7   \\
                       & \textbf{LoRA-SP} & 9.2  & 76  & 73.3 & 80.0 & 80.0 & 80.0 & 60.0 & 80.0 & 93.3 & 80.0    \\
\midrule

\multirow{7}{*}{SmolVLA ~\cite{shukor2025smolvlavisionlanguageactionmodelaffordable} }
                       & \multirow{5}{*}{LoRA  ~\cite{Hu2022LoRA}}   &  1.3 & 8 & 0.0 & 6.7 &  26.7 & 20.0 & 0.0 &  0.0 &  26.7 &  0.0  \\
                       &    &  2.5  & 16     & 53.3 & 60.0 & 80.0 & 60.0 & 0.0 & 13.3 & 86.7 & 0.0   \\
                       &    &  4.9  & 32     & 60.0 & 93.3 & 93.3 & 80.0 & 13.3 & 0.0 & 86.7 & 26.7   \\
                       &    &  9.3  & 64     & 73.3 & 86.7 & 100.0 & 80.0 & 26.7 & 26.7 & 80.0 & 86.7   \\
                       &    &  17.0   & 128    & 86.7 & 80.0 & 100.0 & 100.0 & 40.0 & 20.0 & 93.3 & 86.7   \\
                       & Full FT                   & 100.0 & Full & 86.7 & 86.7 & 100.0 & 100.0 & 73.3  & 86.7 & 100.0 & 86.7   \\
                       & \textbf{LoRA-SP}        & 17.1  & 60 & 86.7 & 93.3 & 93.3 & 100.0 & 86.7 & 86.7 & 100.0 & 93.3   \\
\bottomrule
\end{tabular}
\end{adjustbox}
\label{tab:main_lora}
\vspace{-2pt}
\end{table*}


\vspace{3pt}
\noindent\textbf{Active Vector Selection}
Following Sec.~\ref{sec:analysis}, we determine the active rank by controlling the Frobenius error of low-rank approximation. 
Specifically, we sort generated singular values by $s_i(x)^2$ and define the cumulative energy
\begin{equation}
E_k(x)=\frac{\sum_{i=1}^{k}s_i(x)^2}{\sum_{j=1}^{r}s_j(x)^2}.
\end{equation}
The effective rank $k$ is chosen as the smallest index satisfying $E_k(x)\ge\eta$, and the singular values beyond $k$ are zeroed.
Because $E_k(x)$ bounds the relative approximation error as $\sqrt{1-E_k(x)}$ (Eq.~\ref{eq:relerr}), $\eta$ serves as an explicit tolerance knob (e.g., $\eta=0.99$ implies $\leq 0.1$ error). 
This procedure guarantees that updates retain task-relevant directions while automatically discarding low-energy vectors, yielding efficient and input-specific capacity allocation.

\subsection{Prune: Active-set Reduction via Spectral Loss}\label{sec:prune}


\noindent\textbf{Spectral Loss}
We add a spectral loss term which encourages the router to concentrate energy onto the vectors that survive the $\eta$-based gating:
\begin{equation}
\mathcal{L}_{\text{spec}}(x)=1-E_k(x).
\end{equation}
This creates a reinforcement loop: once a vector is selected, $\mathcal{L}_{\text{spec}}$ pushes its singular value higher, making it even more likely to be selected again. 
Over training, singular-value mass is gradually shifted toward a small stable set of directions, while the task loss prevents collapse to trivial solutions. 
Empirically, we observe that task success rates remain stable even when the number of active vectors is nearly halved, showing that pruning through spectral concentration yields compact adapters with minimal accuracy loss.

\vspace{3pt}
\subsection{Training Losses}\label{sec:loss}
The overall training loss combines task optimization, spectral concentration, and router regularization:
\begin{equation}
\mathcal{L} \;=\; \mathbb{E}[\mathcal{L}_{\text{task}}] \;+\; 10^{-2}\,\mathbb{E}[\mathcal{L}_{\text{spec}}] \;+\; 10^{-3}\,\mathbb{E}[\mathcal{L}_{\text{router}}],
\end{equation}
where $\mathcal{L}_{\text{task}}$ is the main objective (e.g., flow matching), and $\mathcal{L}_{\text{router}}$ includes balance~\cite{fedus2021switch} and $z$-loss~\cite{zoph2022st-moe} terms. 


\section{EXPERIMENTS}
\label{sec:experiments}

\subsection{Experiments Setup}
\label{sec: setup}
\noindent\textbf{Baselines}
We evaluate our method against several widely used and recent fine-tuning strategies, including full fine-tuning (Full FT), standard low-rank adaptation (LoRA), and expressive variants such as LoRA-MoE (top-1 and weighted-sum routing) and AdaLoRA. All methods are trained under identical conditions to ensure fair comparison and our method, LoRA-SP, is trained with an initial rank of 128 and the energy target $\eta=0.9$. These strategies are applied to two pretrained VLA backbones: $\pi_0$, a large-capacity policy based on PaLIGemma~\cite{beyer2024paligemmaversatile3bvlm}, and SmolVLA, an efficient variant built on SmolVLM-2~\cite{marafioti2025smolvlmredefiningsmallefficient}. Performance is evaluated in terms of task success rate under single-task and multi-task training regimes. 

\begin{figure*}[t!]
\centering
\includegraphics[width=\textwidth]{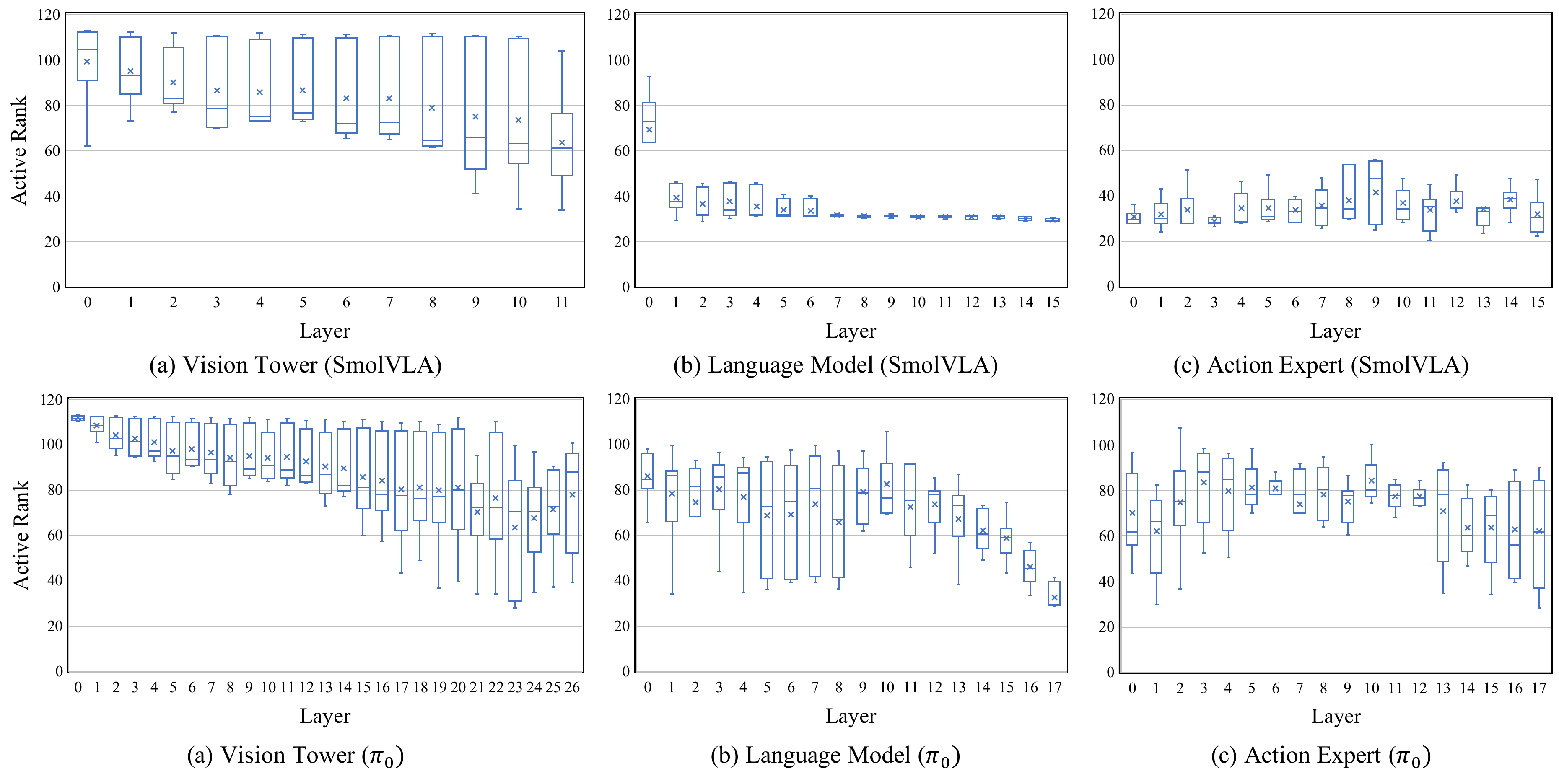}
\caption{
    Layer-wise distributions of active rank learned by LoRA-SP on validation data. 
    Results are shown for (a) vision tower, (b) language model, and (c) action expert, for both SmolVLA (top) and $\pi_{0}$ (bottom) backbones. 
    Each boxplot shows min–LQ–median–UQ–max of the active rank across inputs. 
    The vision tower consistently requires the highest ranks, the action expert shows wide variability, while the language model layers remain comparatively low and stable. 
    This pattern highlights strong heterogeneity across modules and underscores the limitation of using a single global rank for VLA adaptation.
    }
\label{fig:rank_dist_layer}
\vspace{-0.5cm}
\end{figure*}
\begin{figure}[t!]
\centering
\includegraphics[width=0.85\linewidth]{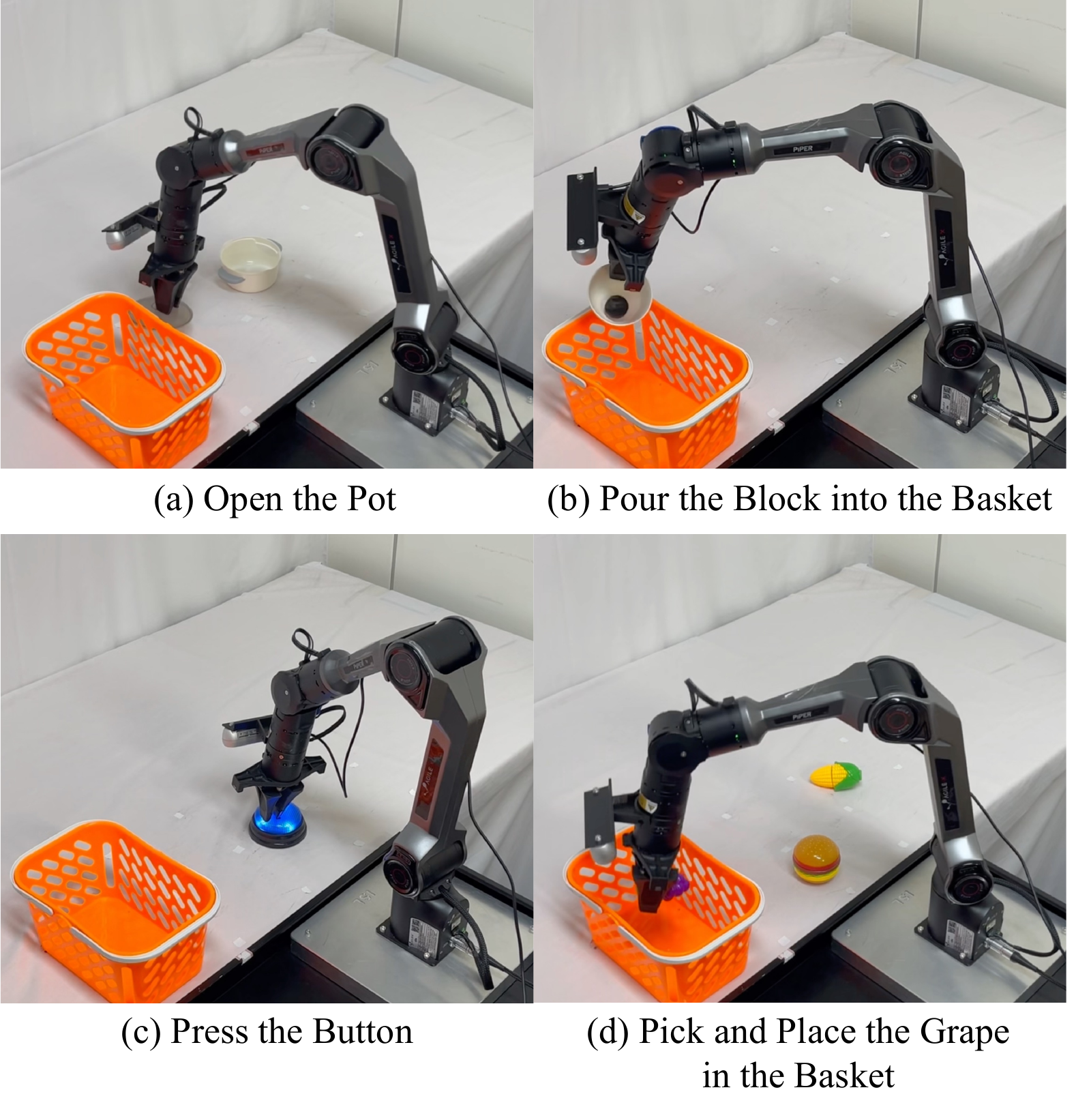}
\caption{Real-world experimental setup for four tasks: (a) Open the Pot, (b) Pour the Block into the Basket, (c) Press the Button, and (d) Pick and Place the Grape in the Basket.}
\label{fig:setup}
\vspace{-2pt}
\end{figure}

\vspace{3pt}
\noindent\textbf{Tasks}
Fig.~\ref{fig:setup} shows the real-world setup for the four manipulation tasks used in our experiments: Open the Pot (Open), Pour the Block into the Basket (Pour), Press the Button (Press), and Pick and Place the Grape in the Basket (Pick-Place). These tasks cover diverse object interactions and assess the model’s generalization and adaptation.

\vspace{3pt}
\noindent\textbf{Robot Embodiment}
All experiments were conducted using the AgileX PiPER robotic arm, a 7-degree-of-freedom (7-DoF) manipulator. This robot is not included in most existing VLA datasets used for pretraining large models, which makes it a suitable testbed for evaluating generalization to unseen embodiments.

\vspace{3pt}
\noindent\textbf{Datasets}
We collected real-world demonstration data using human teleoperation. For each task, we captured 120 episodes, resulting in a total of 480 demonstrations. 
Each episode was collected using two RGB camera views: a side-view camera and a wrist-mounted camera on the robot arm. 

\begin{table}[]
\caption{Ablation study on the spectral loss in LoRA-SP.}
\centering
\setlength\heavyrulewidth{1.2pt}
\begin{adjustbox}{width=0.9\columnwidth}
\begin{tabular}{cccccc}
\toprule
\multirow{2}{*}{} & \multirow{2}{*}{\shortstack{\rule{0pt}{2.5ex} Active Rank \\ (V, L, A)}}  & \multicolumn{4}{c}{Multi-task Success Rate (\%)} \\
 \cmidrule(l){3-6}
   &  & Open & Pour & Press & Pick-Place \\

 \midrule
 \midrule
w/o Spectral    & 83, 107, 57 & 73.3 & 66.7 & 100.0 & 73.3 \\
\textbf{LoRA-SP}  & 84, 35, 34 & 86.7 & 86.7 & 100.0 & 93.3 \\
\bottomrule
\end{tabular}
\end{adjustbox}
\label{tab:ablation_loss}
\vspace{-2pt}
\end{table}
\begin{table}[]
\caption{Ablation study on the energy target $\eta$.}
\centering
\setlength\heavyrulewidth{1.2pt}
\begin{adjustbox}{width=0.9\columnwidth}
\begin{tabular}{cccccc}
\toprule
\multirow{2}{*}{$\eta$} & \multirow{2}{*}{Active Rank} & \multicolumn{4}{c}{Multi-task Success Rate (\%)} \\
 \cmidrule(l){3-6}
   & & Open & Pour & Press & Pick-Place \\

 \midrule
 \midrule
0.5  & 30 & 6.7 & 13.3 & 93.3 & 13.3 \\
0.7  & 46 & 53.3 & 80.0 & 100.0 & 53.3 \\
0.8  & 56 & 80.0 & 86.7 & 100.0 & 60.0 \\
\textbf{0.9}  & 60 & 86.7 & 86.7 & 100.0 & 93.3 \\
0.99 & 114 & 80.0 & 86.7 & 100.0 & 100.0 \\
\bottomrule
\end{tabular}
\end{adjustbox}
\vspace{-2pt}
\label{tab:ablation_threshold}
\end{table}
\subsection{Main Results}
\label{sec: results}
We compare adaptation methods under the multi-task training setting (Table~\ref{tab:main}).
Full fine-tuning yields the highest success rates but comes with high computational cost during training.
Standard LoRA shows degraded performance even at rank 128, particularly on Pour. 
LoRA-MoE and AdaLoRA also fail to consistently outperform standard LoRA. LoRA-MoE lacks fine-grained rank control due to its expert-level gating, while AdaLoRA relies on LLM-based importance scores that misalign with VLA-specific adaptation needs.
Conversely, LoRA-SP sustains uniformly high performance across tasks while modifying relatively few parameters.
It improves average success rates over standard LoRA by 23.3\% on $\pi_0$ and 31.6\% on SmolVLA, while often matching the performance of full fine-tuning.

Table~\ref{tab:main_lora} further compares LoRA across various ranks under both single- and multi-task settings. 
While the success rate in single-task training improves with the rank, that in multi-task training collapses regardless of ranks. 
This is due to two factors: (1) the optimal rank differs by task in terms of its success rate saturation, showing the inadequacy of a single global rank; (2) LoRA lacks task-specific separation, sharing the same low-rank directions across tasks, leading to interference and degraded performance.

To analyze how LoRA-SP allocates capacity, we examine its layer-wise rank distribution across model components. 
Fig.~\ref{fig:rank_dist_layer} shows that the vision module requires consistently high-rank updates, while the language and action modules remain low-rank. 
This highlights a key limitation of fixed-rank methods, which assign uniform capacity regardless of modules. 
In contrast, LoRA-SP adaptively concentrates rank in high-demand modules and prunes others, preserving performance while reducing trainable parameters.

\subsection{Ablation Study}
\label{sec: ablation}
\noindent\textbf{Spectral Loss} 
This ablation compares multi-task performance and module-wise active rank (vision tower, language model, and action expert) with and without the spectral loss. As shown in Table~\ref{tab:ablation_loss}, removing the spectral loss significantly increases active rank, especially in the language module where the active rank rises from 35 to 107. At the same time, task success rates drop across several tasks. The spectral loss guides the model to retain only the most salient rank components, preventing task-irrelevant activations that lead to interference and hinder task success. It preserves higher capacity in the vision tower while pruning redundant ranks in the language and action modules, resulting in improved efficiency without compromising performance.

\vspace{3pt}
\noindent\textbf{Energy target} 
Table~\ref{tab:ablation_threshold} reports an ablation on the energy target $\eta$, which controls the cumulative singular-value energy required to activate rank $k$. 
As $\eta$ decreases, the active rank decreases, reducing the number of basis vectors used in inference. 
We observe a clear trade-off: very low targets (e.g., $\eta=0.5$) severely underfit most tasks, while moderate targets ($\eta=0.7$, $0.8$) achieve strong multi-task performance with substantially fewer active vectors. 
Also, performance saturates around $\eta=0.9$, and setting $\eta=0.99$ nearly doubles the effective rank with marginal additional gains.
This validates our design: the energy target $\eta$ directly tunes the accuracy–efficiency balance, and LoRA-SP maintains high success rates even at reduced ranks.

\section{Conclusion}
We introduced LoRA-SP, a rank-adaptive fine-tuning method for VLA models. Instead of using one fixed rank, LoRA-SP assigns capacity per input and per layer. 
It uses an SVD-style update with vector-level gating, an energy target on the cumulative squared scores. 
This focuses the update on a few useful directions and prunes the rest, yielding compact adapters with robust rank choice and less cross-task interference. 
On four real-robot tasks with two VLA backbones, LoRA-SP matches or exceeds full fine-tuning with far fewer trainable parameters and a smaller active rank, and improves multi-task success by up to 31.6\%.

\bibliographystyle{IEEEtran}
\bibliography{reference}

@article{brohan2023rt2,
  title={Rt-2: Vision-language-action models transfer web knowledge to robotic control},
  author={Zitkovich, Brianna and others},
  booktitle={Conference on Robot Learning},
  pages={2165--2183},
  year={2023},
  organization={PMLR}
}

@article{kim2024openvla,
  title={Openvla: An open-source vision-language-action model},
  author={Kim, Moo Jin and others},
  journal={arXiv preprint arXiv:2406.09246},
  year={2024}
}

@misc{black2024pi0,
  title={$\\pi_0$: A Vision-Language-Action Flow Model for General Robot Control},
  author={Black, Kevin and others},
  year={2024},
  eprint={2410.24164},
  archivePrefix={arXiv},
  primaryClass={cs.LG},
  url={https://arxiv.org/abs/2410.24164}
}

@inproceedings{dou2024loramoe,
  title={LoRAMoE: Alleviating world knowledge forgetting in large language models via MoE-style plugin},
  author={Dou, Shihan and others},
  booktitle={Proceedings of the 62nd Annual Meeting of the Association for Computational Linguistics (Volume 1: Long Papers)},
  pages={1932--1945},
  year={2024}
}

@inproceedings{Li2018IntrinsicDimension,
  author    = {Li, Chunyuan and others},
  title     = {Measuring the Intrinsic Dimension of Objective Landscapes},
  booktitle = {6th International Conference on Learning Representations (ICLR)},
  address   = {Vancouver, BC, Canada},
  year      = {2018}
}

@inproceedings{Aghajanyan2021IntrinsicDim,
  title     = {Intrinsic Dimensionality Explains the Effectiveness of Language Model Fine-Tuning},
  author    = {Aghajanyan, Armen and others},
  booktitle = {Proceedings of the 59th Annual Meeting of the Association for Computational Linguistics and the 11th International Joint Conference on Natural Language Processing (Volume 1: Long Papers)},
  year      = {2021},
  month     = aug,
  address   = {Online},
  publisher = {Association for Computational Linguistics},
  pages     = {7319--7328},
  doi       = {10.18653/v1/2021.acl-long.568}
}

@inproceedings{Hu2022LoRA,
  author    = {Hu, Edward J. and others},
  title     = {LoRA: Low-Rank Adaptation of Large Language Models},
  booktitle = {The Tenth International Conference on Learning Representations (ICLR 2022)},
  year      = {2022},
  publisher = {OpenReview.net}
}

@article{Zhang2023AdaLoRA,
  author   = {Zhang, Qingru and others},
  title    = {Adaptive Budget Allocation for Parameter-Efficient Fine-Tuning},
  journal  = {CoRR},
  volume   = {abs/2303.10512},
  year     = {2023},
  doi      = {10.48550/arXiv.2303.10512},
  eprint   = {2303.10512},
  eprinttype = {arXiv}
}

@article{deepmind2025geminirobotics,
  title={Gemini Robotics: Bringing AI into the Physical World},
  author={Google DeepMind},
  journal={arXiv preprint arXiv:2503.20020},
  year={2025},
  eprint={2503.20020},
  archivePrefix={arXiv}
}

@article{brohan2022rt1,
  title={RT-1: Robotics Transformer for Real-World Control at Scale},
  author={Brohan, Anthony and others},
  journal={arXiv preprint arXiv:2212.06817},
  year={2022},
  eprint={2212.06817},
  archivePrefix={arXiv},
  primaryClass={cs.RO}
}

@inproceedings{oxe2023rtx,
  title={Open x-embodiment: Robotic learning datasets and rt-x models: Open x-embodiment collaboration 0},
  author={O’Neill, Abby and others},
  booktitle={2024 IEEE International Conference on Robotics and Automation (ICRA)},
  pages={6892--6903},
  year={2024},
  organization={IEEE}
}

@article{dettmers2023qlora,
  title={Qlora: Efficient finetuning of quantized llms},
  author={Dettmers, Tim and others},
  journal={Advances in neural information processing systems},
  volume={36},
  pages={10088--10115},
  year={2023}
}

@inproceedings{liu2024dora,
  title={Dora: Weight-decomposed low-rank adaptation},
  author={Liu, Shih-Yang and others},
  booktitle={Forty-first International Conference on Machine Learning}
}

@article{fedus2021switch,
  title={Switch transformers: Scaling to trillion parameter models with simple and efficient sparsity},
  author={Fedus, William and others},
  journal={Journal of Machine Learning Research},
  volume={23},
  number={120},
  pages={1--39},
  year={2022}
}

@article{levine2016e2e,
  title   = {End-to-End Training of Deep Visuomotor Policies},
  author  = {Levine, Sergey and others},
  journal = {Journal of Machine Learning Research},
  volume  = {17},
  number  = {39},
  pages   = {1--40},
  year    = {2016},
  url     = {https://jmlr.org/papers/v17/15-522.html}
}

@inproceedings{rajeswaran2018dexterous,
  title     = {Learning Complex Dexterous Manipulation with Deep Reinforcement Learning and Demonstrations},
  author    = {Rajeswaran, Aravind and others},
  booktitle = {Robotics: Science and Systems (RSS)},
  year      = {2018},
  doi       = {10.48550/arXiv.1709.10087},
  url       = {https://arxiv.org/abs/1709.10087}
}

@inproceedings{kalashnikov2018qtopt,
  title     = {Scalable Deep Reinforcement Learning for Vision-Based Robotic Manipulation},
  author    = {Kalashnikov, Dmitry and others},
  booktitle = {Proceedings of the 2nd Conference on Robot Learning (CoRL)},
  series    = {Proceedings of Machine Learning Research},
  volume    = {87},
  year      = {2018},
  url       = {https://proceedings.mlr.press/v87/kalashnikov18a/kalashnikov18a.pdf}
}

@misc{shukor2025smolvlavisionlanguageactionmodelaffordable,
  title={SmolVLA: A Vision-Language-Action Model for Affordable and Efficient Robotics},
  author={Shukor, Mustafa and others},
  year={2025},
  eprint={2506.01844},
  archivePrefix={arXiv},
  primaryClass={cs.LG},
  url={https://arxiv.org/abs/2506.01844}
}

@misc{beyer2024paligemmaversatile3bvlm,
  title={PaliGemma: A versatile 3B VLM for transfer},
  author={Beyer, Lucas and others},
  year={2024},
  eprint={2407.07726},
  archivePrefix={arXiv},
  primaryClass={cs.CV},
  url={https://arxiv.org/abs/2407.07726}
}

@misc{marafioti2025smolvlmredefiningsmallefficient,
  title={SmolVLM: Redefining small and efficient multimodal models},
  author={Marafioti, Andrés and others},
  year={2025},
  eprint={2504.05299},
  archivePrefix={arXiv},
  primaryClass={cs.AI},
  url={https://arxiv.org/abs/2504.05299}
}

@article{openai2023gpt4,
  title   = {GPT-4 Technical Report},
  author  = {{OpenAI}},
  journal = {arXiv preprint arXiv:2303.08774},
  year    = {2023},
  url     = {https://arxiv.org/abs/2303.08774}
}

@article{gemini2023,
  title   = {Gemini: A Family of Highly Capable Multimodal Models},
  author  = {{Gemini Team} and others},
  journal = {arXiv preprint arXiv:2312.11805},
  year    = {2023},
  url     = {https://arxiv.org/abs/2312.11805}
}

@article{llava2023,
  title={Visual instruction tuning},
  author={Liu, Haotian and others},
  journal={Advances in neural information processing systems},
  volume={36},
  pages={34892--34916},
  year={2023}
}

@article{touvron2023llama,
  title={Llama: Open and efficient foundation language models},
  author={Touvron, Hugo and others},
  journal={arXiv preprint arXiv:2302.13971},
  year={2023}
}

@article{zoph2022st-moe,
  title={St-moe: Designing stable and transferable sparse expert models},
  author={Zoph, Barret and others},
  journal={arXiv preprint arXiv:2202.08906},
  year={2022}
}

\end{document}